\documentclass{article}

\usepackage{PRIMEarxiv}

\usepackage[utf8]{inputenc} 
\usepackage[T1]{fontenc}    
\usepackage{hyperref}       
\usepackage{url}            
\usepackage{booktabs}       
\usepackage{amsfonts}       
\usepackage{nicefrac}       
\usepackage{microtype}      
\usepackage{lipsum}
\usepackage{xcolor}
\usepackage{fancyhdr}       
\usepackage{graphicx}       
\graphicspath{{media/}}     

\usepackage{amsmath} 
\usepackage{amsthm} 
\usepackage{amstext}
\usepackage{subfigure}
\usepackage{authblk}
\usepackage{hyperref}

\title{Deciphering the Projection Head: Representation Evaluation Self-supervised Learning
}

\author[*]{Jiajun Ma}
\author[**]{Tianyang Hu}
\author[*]{Wenjia Wang}
\affil[*]{Hongkong University of Science and Technology}
\affil[**]{Huawei Noah’s Ark Lab}
\affil[ ]{\texttt{jmabh@connect.ust.hk}, 
\texttt{hutianyang1@huawei.com}, 
\texttt{wenjiawang@ust.hk}}

\begin{document}
\maketitle

\begin{abstract}
Self-supervised learning (SSL) aims to learn intrinsic features without labels. Despite the diverse architectures of SSL methods, the projection head always plays an important role in improving the performance of the downstream task. In this work, we systematically investigate the role of the projection head in SSL. Specifically, the projection head targets the uniformity part of SSL, which pushes the dissimilar samples away from each other, thus enabling the encoder to focus on extracting semantic features. Based on this understanding, we propose a Representation Evaluation Design (RED) in SSL models in which a shortcut connection between the representation and the projection vectors is built. Extensive experiments with different architectures, including SimCLR, MoCo-V2, and SimSiam, on various datasets, demonstrate that the representation evaluation design can consistently improve the baseline models in the downstream tasks. The learned representation from the RED-SSL models shows superior robustness to unseen augmentations and out-of-distribution data.

\end{abstract}


\section{Introduction}
\label{Introduction}
Extracting meaningful representations from a large amount of unlabeled data is an important task in self-supervised learning.
With the rapid progress in the SSL models \cite{chen2020simple,he2020momentum,grill2020bootstrap,chen2021exploring,he2022masked}, a simple classifier learned from the pre-trained representations can achieve comparable performance to direct supervised learning. 

Although SSL has achieved great empirical success, the intrinsic understanding of the mechanism behind it still needs to be explored. Many efforts have been devoted to studying the loss function, and the construction of positive pairs \cite{arora2019theoretical,wen2021toward,wang2020understanding,tian2021understanding,wang2021understanding,wang2022chaos}, while less are paid on the investigation of the architectures. 
Typically, the architecture of the SSL method includes two parts: an encoder and a projection head. The encoder is usually a Res-Net \cite{he2016deep} that aims to extract semantic features, and the projection head is a multi-layer perceptron used in pre-training loss calculation. After the pre-training, the projection head is discarded, and the encoder outputs are used for the downstream tasks (usually classification tasks). It has been shown that the projection head can significantly improve the performance of SSL methods \cite{chen2020simple}; thus, the projection head design has been widely adopted in diverse SSL models \cite{chen2020simple,he2020momentum,chen2020improved,caron2020unsupervised,grill2020bootstrap,zbontar2021barlow,chen2021exploring,he2022masked}. However,
the role of the projection head has yet to be identified. Therefore, a deeper understanding of the projection head is called for. 

\begin{figure}[h]
  \centering
    \includegraphics[width=0.5\linewidth]{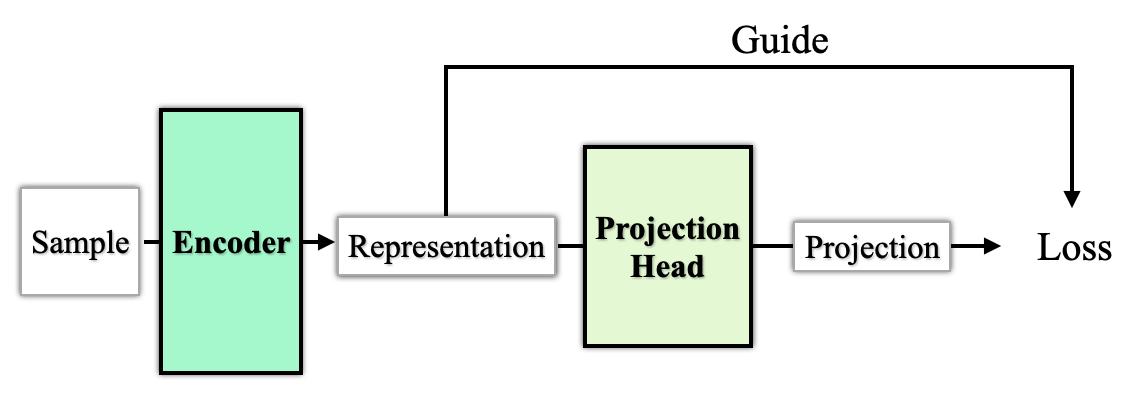}
  \caption{\textbf{RED-SSL architecture.}}
  \label{fig:red_idea}
\end{figure}

In this work, with a thorough investigation of the projection head, we display that the projection head focuses on the uniformity objective, maximizing the distance between dissimilar samples. Meanwhile, the encoder pays more attention to enhancing alignment, minimizing the distance between similar samples (e.g., a training data point and its augmentations). We demonstrate that this phenomenon widely exists in SSL methods, including contrastive methods such as SimCLR and MoCo-V2, and non-contrastive methods such as SimSiam. Despite the alignment being more related to extracting semantic features, the projection head can effectively prevent training collapse.

Based on this deeper understanding of the projection head, we introduce a Representation Evaluation Design in the SSL methods named RED, where a shortcut connection between the representation vectors and the projection vectors is built, illustrated in Figure \ref{fig:red_idea}. During the back-propagation in the training process, the shortcut allows the gradient from the representation layer to bypass the projection head and guide the training directly.  
Through comprehensive comparison experiments in the SimCLR, MoCo-V2, and SimSiam architectures and their corresponding models with RED design (RED-SimCLR, RED-MoCo-V2, and RED-SimSiam) in Cifar10, Cifar100 \cite{Krizhevsky09learningmultiple}, ImageNet1000 \cite{deng2009imagenet} and mixed-Guassian simulated data, we observe a consistent performance improvement in the downstream classification tasks, with different classification methods including k-nearest neighbor (kNN) and linear classifier.
Furthermore, the representations learned from RED-SSL exhibit stronger robustness to augmentation in the downstream evaluation.

Our main contributions are summarized as follows.
\begin{itemize}
\item We uncover that the projection head is a uniform projector, regardless of whether the uniformity part appears explicitly in the objective function of SSL. Thus, the projection head enables the encoder to focus on boosting alignment without worrying about training collapse. It explains the combination of the encoder and projection head outperforms the individual encoder.
 
\item We show the encoder outputs (representation vectors) exhibit superiority in terms of augmentation robustness, lower entropy, and better downstream task performance than the outputs of the projection head (projection vectors). It explains why the representation vectors are used in the downstream tasks.
 
\item We propose a representation evaluation design (RED) that bridges the representation information and the SSL objective functions. Extensive experiments on different SSL methods, various datasets, and different classifiers demonstrate that our proposed design can consistently improve the downstream task performance of the baseline models and is more robust to unseen augmentations and out-of-distribution data.
\end{itemize}

\section{Related Works} 
\begin{figure}[h]
  \centering
    \includegraphics[width=\linewidth]{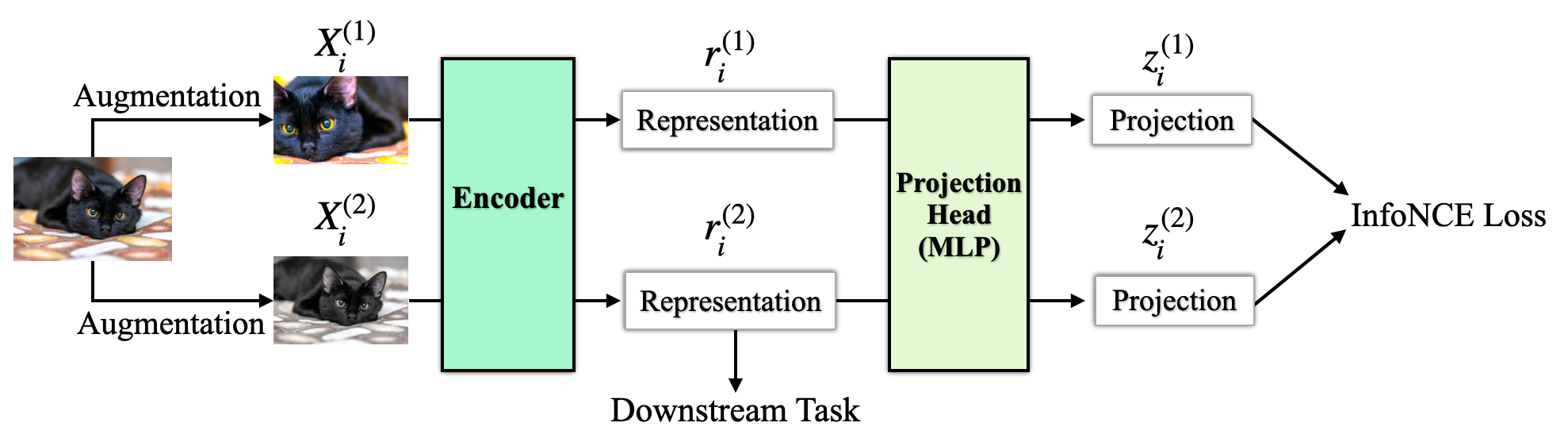}
  \caption{\textbf{SimCLR architecture} \cite{chen2020simple}. The representation $r$ before the projection head is used for downstream tasks, and the projection layer $z$ is used for InfoNCE loss calculation. This projection head design is widely accepted in later diverse contrastive models. }
  \label{fig:SimCLR_projection_head}
\end{figure}

\begin{figure}[h]
  \centering
\includegraphics[width=\linewidth]{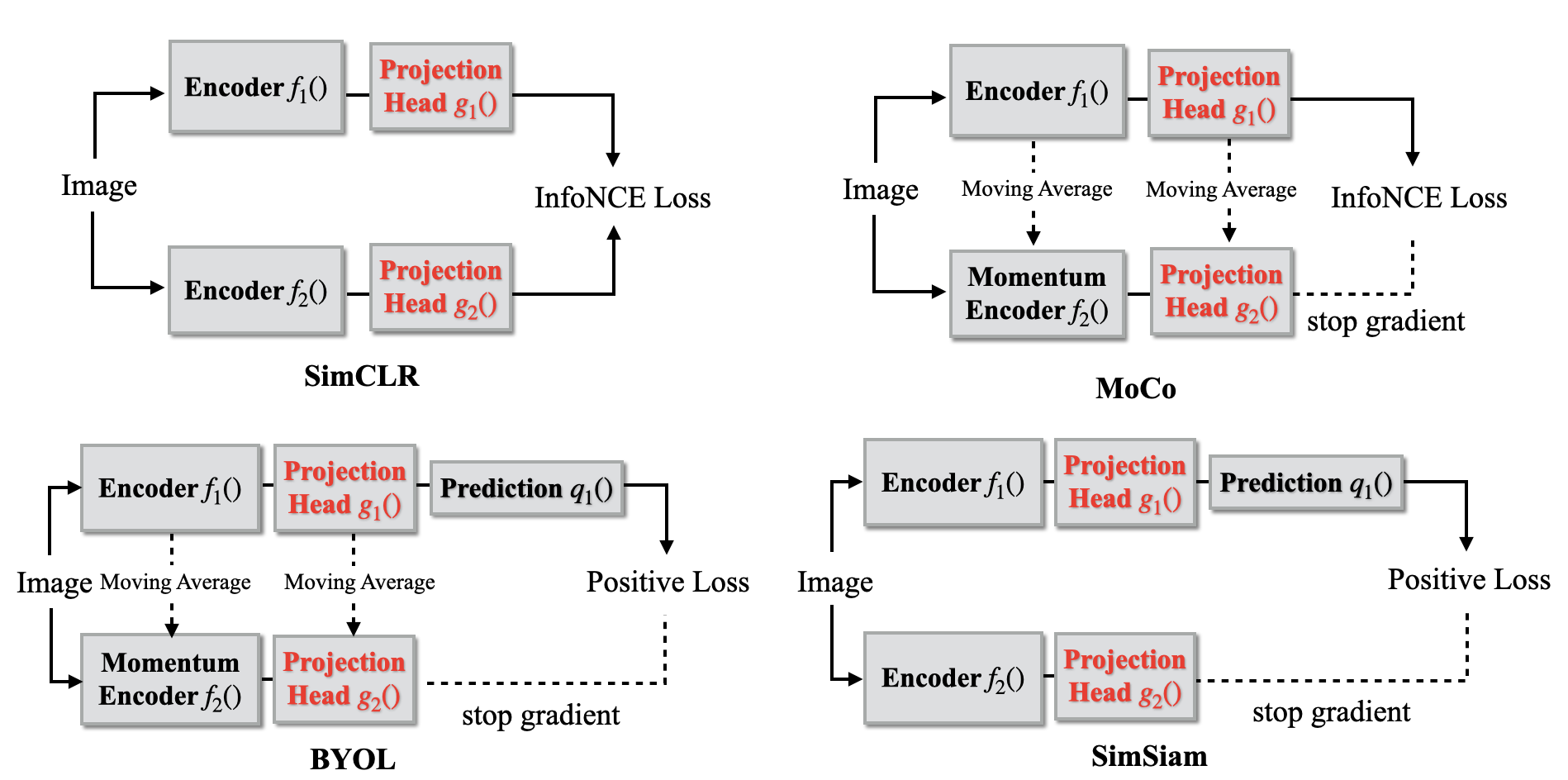}
  \caption{\textbf{Demonstration of different network architectures}, SimCLR, MoCo-V2, BYOL, SimSiam all share the projection head design.}
  \label{fig:Network_architecture}
\end{figure}

\textbf{The projection head in SSL.}
The projection head design was initially introduced in SimCLR \cite{chen2020simple} as displayed in Figure \ref{fig:SimCLR_projection_head}. In SimCLR, the projection head differentiates the pre-training and the downstream usage as separate objectives for the projection vector $z$ and representation vector $r$, respectively. This projection head design is widely adopted by the later proposed methods, including MoCo-V2 \cite{he2020momentum, chen2020improved}, BYOL \cite{grill2020bootstrap}, SwAV
\cite{caron2020unsupervised}, Barlow Twins
\cite{zbontar2021barlow}, SimSiam \cite{chen2021exploring}, MAE
\cite{he2022masked} where the architectures are displayed in 
Figure \ref{fig:Network_architecture}. \cite{chen2020simple} indicates that  the downstream classification does not increase monotonically as the number of layers of projection head increases.
In \cite{wang2022revisiting}, they indicate that the projection head is the key to enhancing the transferability; and with the projection head included, even a supervised model can enjoy transferability gain. 
In \cite{gupta2022understanding}, they regard the projection head as a low-rank mapping such that the trained vectors can be more style-invariant and generalize better. This interpretation may fail to explain the contrastive learning model in \cite{appalaraju2020towards}, where a fixed uniformly distributed projection head (which is not low rank) can still improve the downstream performance. 

\begin{figure*}[t]
    \centering
    \subfigure[]{\includegraphics[width=0.31\textwidth]{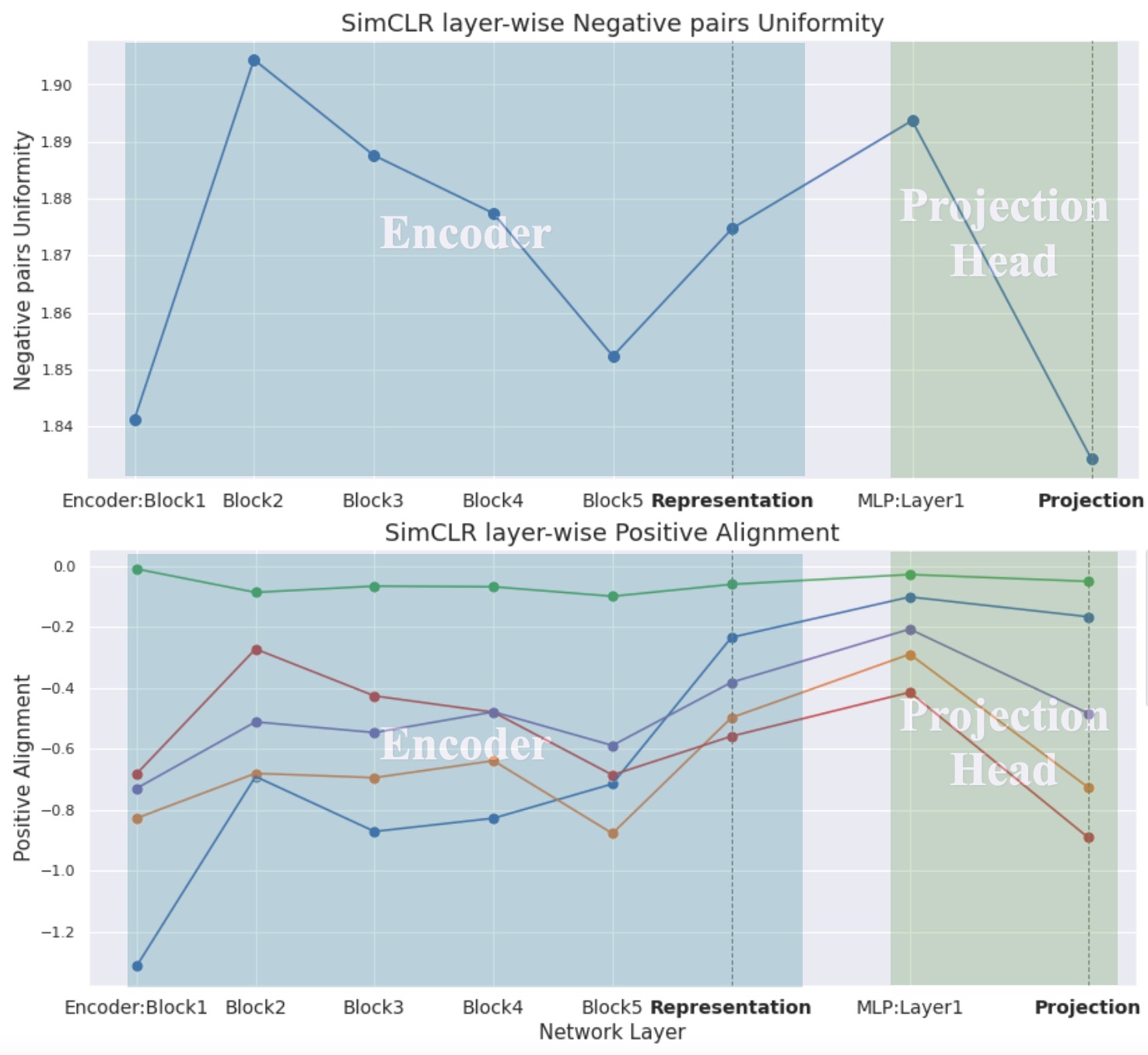}}
    \subfigure[]{\includegraphics[width=0.31\textwidth]{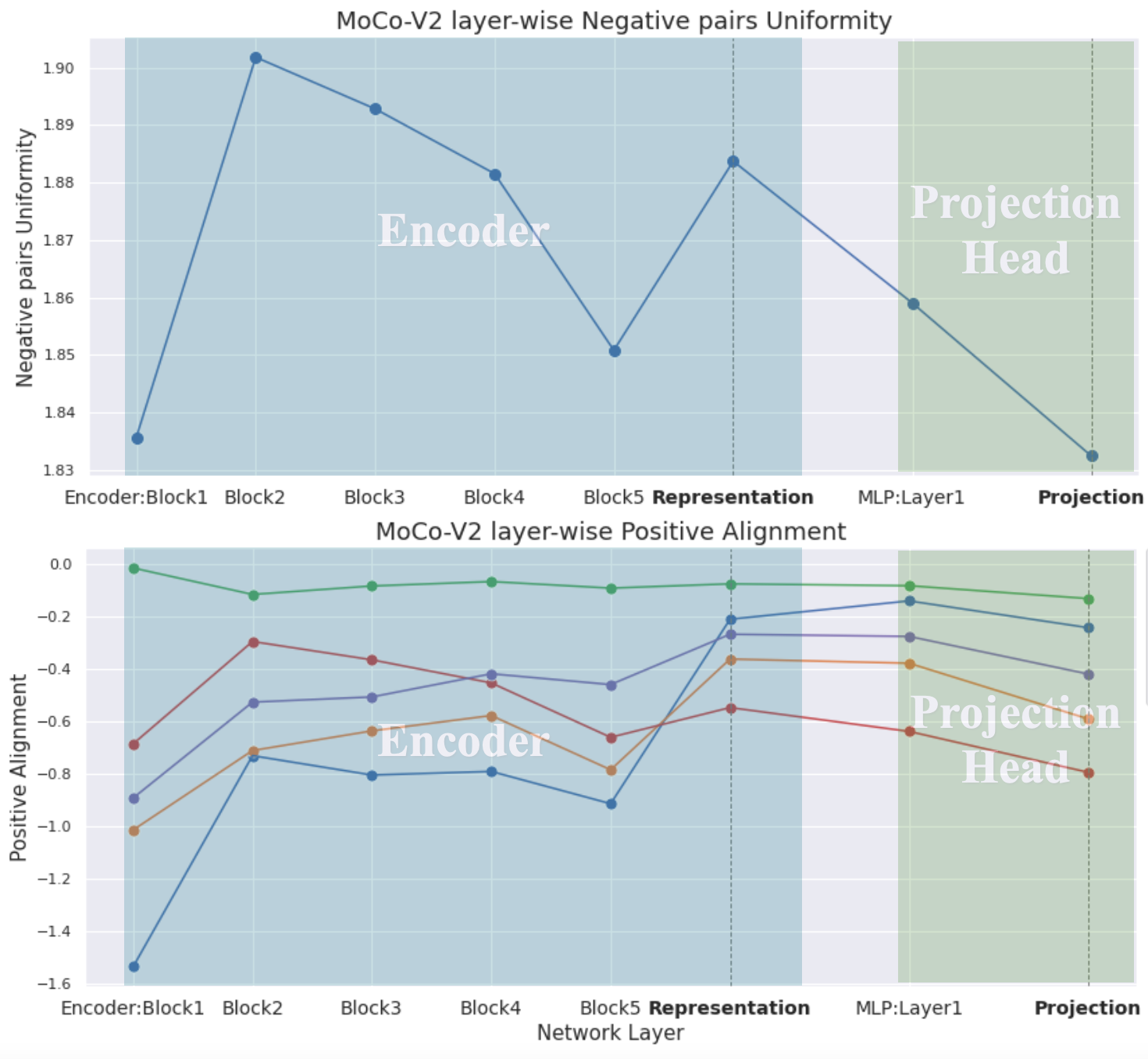}} 
    \subfigure[]{\includegraphics[width=0.37\textwidth]{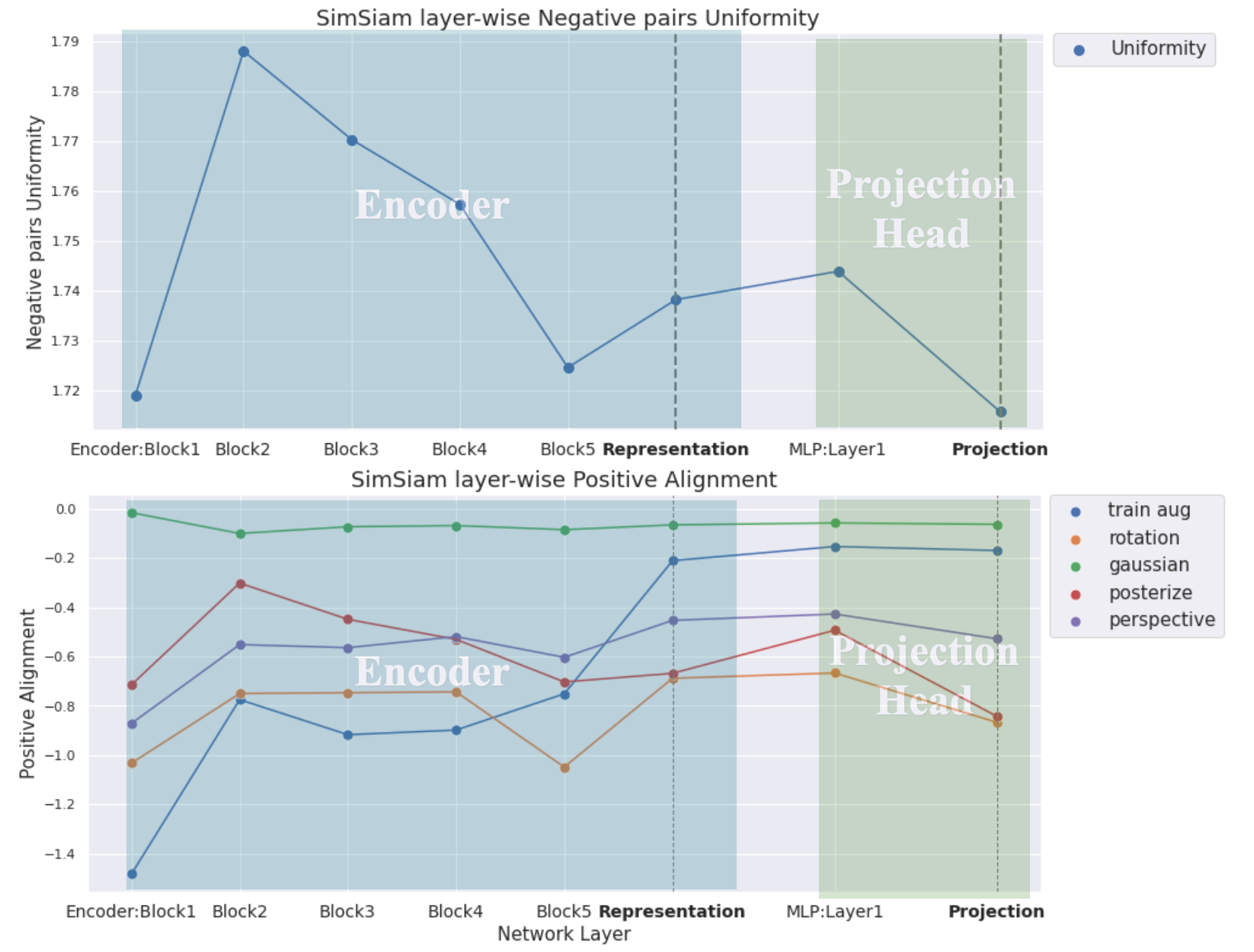}} 
    \caption{\textbf{Layer-wise analysis}. Conducting the alignment and uniformity calculation based on each intermediate layer within the SSL architectures. (a) SimCLR (b) MoCo-V2. (c) SimSiam.  }
    \label{fig:ssl_layer_total}
\end{figure*}

\textbf{Analysis of SSL.}
The superiority of SSL generalization has been the spotlight of many previous works \cite{arora2019theoretical,tosh2021contrastive,haochen2021provable,wen2021toward,ji2021power}. The analysis of the SSL loss and the architecture is relatively fewer.      
\cite{wang2020understanding} split the InfoNCE loss as alignment(cosine similarity between positive pairs) and uniformity(logarithm of the sum of pairwise cosine similarity of negative pairs), and demonstrate the uniformity stands for uniformly distributing on the hypersphere based on Gaussian potential kernel.
\cite{hu2022your} relates contrastive SSL to the stochastic neighbor embedding(SNE).
\cite{wang2022chaos} considers augmentation encourages the different intra-class samples to be overlapped, and thus positive alignment could attract the intra-class samples together.
\cite{wang2021understanding} states that uniformity guides the contrastive model to learn separable features, and proper temperature gives tolerance to semantically similar samples. 
\cite{wen2022mechanism} reveals that the identity-initialized prediction head prevents BYOL from the training collapse. 
\cite{saunshi2022understanding}
point out that the inductive biases within the contrastive function class contribute to the downstream success.

\section{Projection Head is Uniformity Projector}
In this section, we uncover the encoder and projection head intrinsically focus on different parts of the SSL objective. Specifically,
After investigating the layer-wise alignment \& uniformity, we reveal that the projection head, in essence, targets the uniformity objective. Therefore, with the projection head, SSL can enable the encoder to spotlight boosting alignment without worrying about uniformity. 

In one of the most popular SSL methods, SimCLR, the objective function (called InfoNCE loss) is \cite{chen2020simple,yeh2022decoupled}
\begin{equation}
\label{eqn:loss_alignment_uniformity}
\begin{aligned}
 &  \sum_{i=1}^n l_{i}=- \sum_{i=1}^n\log \frac{ \exp(z^{(1)}_i z^{(2)}_i / \tau )}{\sum_{j\in \{1,n\}, k,l\in \{1,2\}} \exp(z^{(k)}_i z^{(l)}_j / \tau) } \\
      & = - \sum_{i=1}^n (z^{(1)}_i z^{(2)}_i / \tau) + \sum_{i=1}^n \log \sum_{j=1}^n \exp(z^{(k)}_i z^{(l)}_j / \tau) \\
      & = - \text{alignment} + \text{uniformity} \\ 
 \end{aligned}
\end{equation}
where $n$ is the batch size, $z_i$ is the projection vector of a sample $i$, $n$ is the batch size. Referring the same sample under different augmentation $z^{(1)}_i z^{(2)}_i$ as positive pairs, pairwise sample under different augmentation $z^{(k)}_i z^{(l)}_j$ as negative pairs. Superscript ${(1)}, {(2)}$ of $z_i$ indicates the augmentation for the positive pairs of projection $i$;  superscript $(k), (l)$ indicates the augmentation for the negative pairs of projection $i, j$. In the InfoNCE loss, the first term represents the negative alignment. Alignment is maximized if the projection $z$ is invariant to the training augmentation. The second term represents uniformity, which is minimized if the projection $z$ is mapped as the uniform distribution on the hype-sphere. Motivated by the InfoNCE loss, we calculate the alignment and uniformity by the first and second terms in \eqref{eqn:loss_alignment_uniformity}, respectively, for different methods, including SimCLR, MoCo-V2, and SimSiam.

Figures \ref{fig:ssl_layer_total} depict the dynamics of alignment calculation based on each intermediate layer within the SSL architectures. (a) SimCLR (b) MoCo-V2. (c) SimSiam; all are trained for 200 epochs in Cifar10, and the encoder is ResNet-18.
From these figures, it can be seen that the uniformity  
shakes within the encoder part; as stepping into the projection head, an obvious decrease in uniformity occurs. The decrease in the uniformity coincides with the alignment decrease in the projection head, 
indicating that the samples are mapped closer to the uniform distribution on the hyper-sphere at the cost of impairing positive alignment. Table \ref{table:changes_alignment_uniformity} explicitly lists the changes of uniformity and alignment within the encoder and projection head. The encoder focuses on boosting the alignment and pays less attention to decreasing uniformity. For projection head, it acts as the uniformity projector: decreasing the uniformity at the cost of reducing alignment. Thus, with the projection head targeting the uniformity objective, the encoder can concentrate on promoting alignment and extracting semantically meaningful features. This
explains why introducing the projection head can improve the downstream performance of the encoder output (representation).

\begin{table}[ht]
\caption{\textbf{Uniformity and alignment changes within the encoder and projection head}. The numbers in the second column are obtained by (the uniformity of the last layer) $-$ (the uniformity of the first layer). The numbers in the third column are obtained similarly for alignment.}
\label{table:changes_alignment_uniformity}
\vskip 0.15in
\begin{center}
\begin{small}
\begin{sc}
\begin{tabular}{lcc}
\toprule
 SimCLR & Uniformity  & Alignment \\
\midrule
 Encoder   & +0.035 & +0.321 \\
 Projection Head  & -0.057 & -0.212  \\
\midrule
MoCo-V2 & Uniformity  & Alignment \\
\midrule
 Encoder   & +0.049 & +0.481 \\
 Projection Head  & -0.028 & -0.151  \\
\midrule
SimSiam & Uniformity  & Alignment \\
\midrule
 Encoder   & +0.019 & +0.401 \\
 Projection Head  & -0.025 & -0.143  \\
\bottomrule
\end{tabular}
\end{sc}
\end{small}
\end{center}
\vskip -0.1in
\end{table}

It is worth noting that SimSiam does not explicitly require uniformity in the training objective function. The significant decrease of uniformity within the projection head in the upper sub-figure of Figure \ref{fig:ssl_layer_total} (c) suggests that the projection head still implicitly enables uniformity and prevents the training collapse.
It also
explains the experiment in \cite{appalaraju2020towards}, where they introduce a training-frozen projection head with uniformly simulated parameters still enhance the downstream performance of the encoder output. It is because the projection head can still map the sample into uniformity even if the parameters are frozen, thus allowing the encoder to pursue the alignment objective better.

\section{Representation Vector Analysis}
With the projection head fulfilling the uniformity requirement, the encoder can focus on the alignment, and thus intuitively, the representation vectors contain more semantic information. To verify this intuition, we conduct experiments in this section to show that
the representation vectors exhibit superiority in augmentation robustness, lower entropy, and better downstream task performance over the projection vectors. This representation superiority lays the foundation of our proposed representation evaluation design (RED) applicable in SSL models.  

\subsection{Robustness to unseen augmentations}\label{subsec_robust}

The augmentation invariance guides the semantic features extraction in SSL \cite{von2021self, purushwalkam2020demystifying, wang2022chaos}, and demonstrated in the successes of non-contrastive models \cite{grill2020bootstrap, chen2021exploring, he2022masked}. 
During the training process, only several types of augmentation are applied
 (random combination of RandomResizedCrop, HorizontalFlip, ColorJitter, and RandomGrayscale, specified in \cite{chen2020simple}). However,
other unseen augmentations can contribute to semantic information extraction as well. 

In order to study the robustness to unseen augmentations of the representation vectors and projection vectors, we compare the cosine similarity of the positive pairs based on the representation and the projection vectors. Table \ref{table:repr_proj_positive_augmentation} records the cosine similarity between positive sample pairs under different augmentations, calculated with representation and projection vectors, respectively. The train augmentation stands for the augmentation types used during the training (random combination of RandomResizedCrop, HorizontalFlip, ColorJitter, and RandomGrayscale, specified in \cite{chen2020simple}). 
The projection vectors show a higher cosine similarity for the train augmentation, but for those unseen augmentations, 
the cosine similarity of the representation vectors is significantly higher than that of the projection vectors, except for Gaussian blur, which is not a semantic-meaningful augmentation. This superior augmentation robustness indicates that the representation vectors' capacity to extract meaningful semantic information is not limited to the pre-determined augmentation types. 

\begin{table}[h]
\caption{\textbf{Augmentation robustness of representation and projection}. The cosine similarity between positive sample pairs under different types of augmentations. The following abbreviations are used:
TA = Train augmentation stands for the augmentation types used during the pre-training; AR = Angle rotate; GB = Gaussian Blur; CC = Center Crop; RPo = Random Posterize; RPe = Random Perspective. The SSL model is SimCLR trained for 200 epochs in Cifar10 with ResNet-18.}
\label{table:repr_proj_positive_augmentation}
\vskip 0.15in
\begin{center}
\begin{small}
\begin{sc}
\begin{tabular}{lcccc}
\toprule
     & \multicolumn{2}{c}{Representation} & \multicolumn{2}{c}{Projection} \\ 
     &  Train &Test & Train & Test \\
\midrule
TA & 0.823 & 0.802 & 0.894 & 0.857   \\
AR  &  0.559 & 0.558 & 0.479 & 0.470       \\
GB &  0.951 & 0.947 & 0.965 & 0.958      \\
CC	  &  0.219 & 0.221 & 0.134 & 0.130 \\
RPo &  0.509 & 0.509 & 0.422 & 0.413 \\
RPe & 0.730 & 0.728 & 0.704 & 0.698  \\
\bottomrule
\end{tabular}
\end{sc}
\end{small}
\end{center}
\vskip -0.1in
\end{table}

\subsection{Entropy analysis}
Uniformity plays an essential role in avoiding collapse into a trivial constant model \cite{wang2021understanding}. However, encouraging uniformity comes with a byproduct of increasing entropy (see Proposition \ref{prop_u}). Thanks to the projection head design, the encoder transfers the uniformity responsibility to the projection head, thus avoiding increasing the entropy of representation vectors. In this subsection, we reveal that the representation vectors enjoy smaller entropy than the projection vectors. 

\newtheorem{prop}{Proposition}
\begin{prop}\label{prop_u}
Encouraging uniformity is equivalent to reducing the KL divergence towards the uniform distribution, and approaching closer to the uniform distribution results in an entropy increase.
\end{prop}

\begin{figure}[ht]
\vskip 0.2in
\begin{center}
\centerline{\includegraphics[width=0.8\columnwidth]{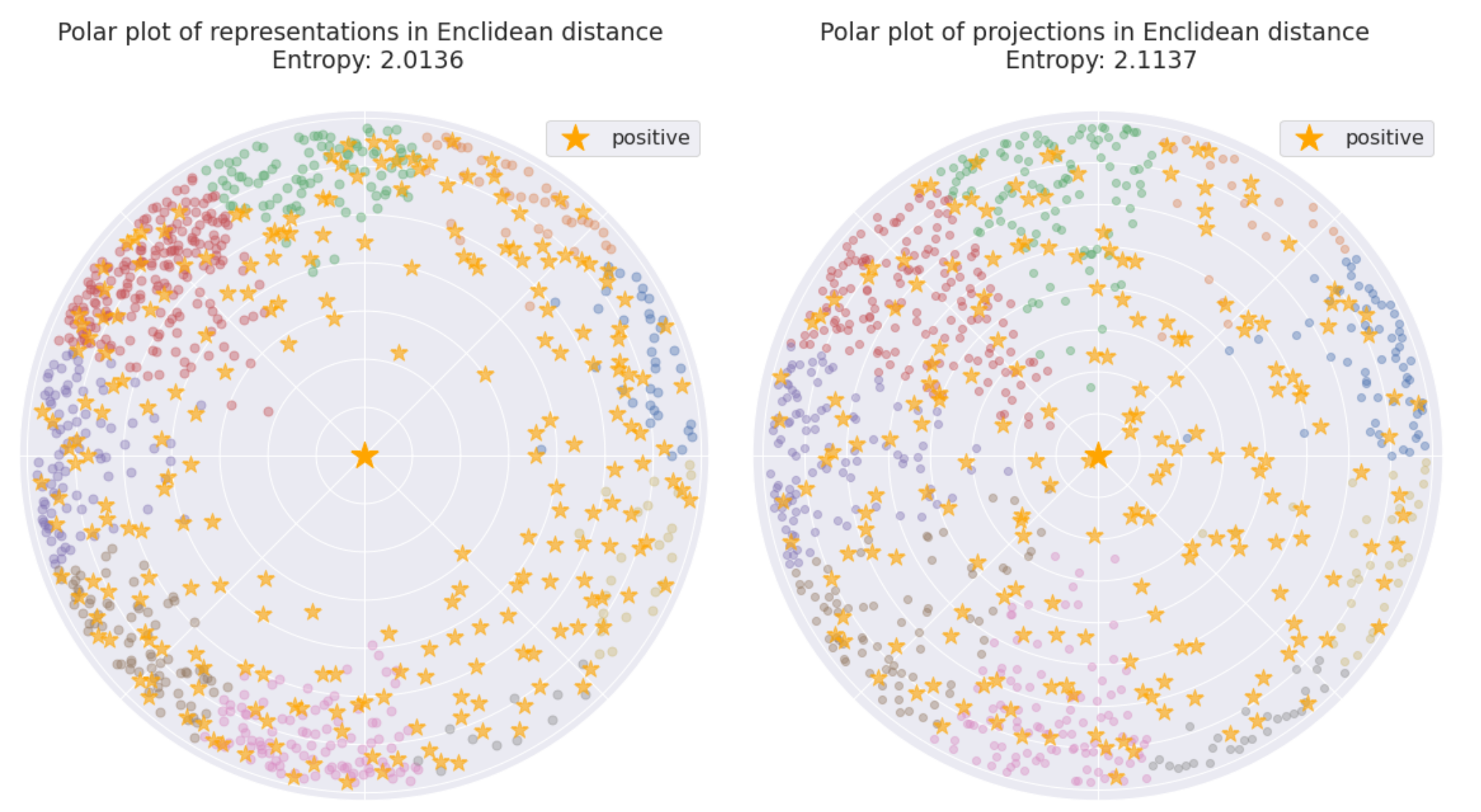}}
\caption{\textbf{Polar plots of entropy demonstration}. The polar plots of samples were measured with representation and projection vectors, respectively. The points are the closest 1000 samples to the origin point. The orange points stand for the sample pairs sharing the same label with the big orange star at the origin point, while the other color points correspond to the samples with different labels with the orange star. The estimated entropy of the left representation-based polar plot is 2.0136, and the right is 2.1137. The SSL model is SimCLR trained for 200 epochs in Cifar10 with ResNet-18 as the encoder.}
\label{fig:repr_proj_polar_plot}
\end{center}
\vskip -0.2in
\end{figure}

The proof of Proposition \ref{prop_u} is based on the derivation in \eqref{eqn:uniformity_probability} and \eqref{eqn:kl_uniformity_probability} of Appendix \ref{appendix:Uniformity_entropy}. Since the uniform distribution is the distribution that has the largest entropy for bounded variables, and the projection head encourages uniformity, the output distribution of the projection head has a higher entropy than its original distribution. This phenomenon is illustrated in Figure \ref{fig:repr_proj_polar_plot}. Figure \ref{fig:repr_proj_polar_plot} depicts the polar plots of sample pairs based on representation and projection vectors separately.
We can see that the negative pairs in projection vectors are more volatile and mixed with positive pairs, which is more likely to result in misclassified sample pairs in the downstream task. To provide a quantitative comparison, we conduct the discrete entropy estimator \cite{beirlant1997nonparametric} defined as 
\begin{align}
\label{eqn:entropy_formula}
& \hat{H} = -\sum_{\text{label}=1}^{k} \hat{p}_{\{\text{label}\}} \log \hat{p}_{\{\text{label}\}},
\end{align}
where
$$ \hat{p}_{\text{\{label = i\}}} = \frac{\text{numbers of label i samples in bandwidth}}{\text{total numbers of samples in bandwidth}}.$$
The estimated entropy of the left representation calculated polar plot is 2.0136, while the right projection calculated polar plot is 2.1137, with the bandwidth chosen as 0.1. 

\begin{table}[!ht]
\caption{\textbf{Entropy in SSL}. Entropy of the representation and projection vectors in SimCLR, MoCo-V2, and SimSiam.}
\label{table:repr_proj_entropy}
\vskip 0.15in
\begin{center}
\begin{small}
\begin{sc}
\begin{tabular}{lcc}
\toprule
Entropy & Representation & Projection \\
\midrule
SimCLR   &  1.7476 & 1.850  \\
MoCo-V2  &  1.5131 & 2.0315   \\
SimSiam  &  1.8611 & 1.8940 \\
\bottomrule
\end{tabular}
\end{sc}
\end{small}
\end{center}
\vskip -0.1in
\end{table}

Table \ref{table:repr_proj_entropy} lists the comparison of the estimated entropy calculated with the representation and projection vectors in SimCLR, MoCo-V2, and SimSiam; all are trained for 200 epochs in Cifar10 with ResNet-18 as the encoder. It can be seen that the estimated entropy calculated with the representation vectors is consistently smaller than that of the projection vectors. It indicates that with the introduction of the projection head, the encoder shifts the uniformity burden to the projection head; thus, the representation vectors enjoy a smaller entropy and contain a wealth of information useful for downstream tasks.     

\subsection{Relationship to downstream task}\label{subsec_relation_task}
With the projection head targeting the uniformity objective, SSL models allow the representation vectors to have higher alignment; thus,
the quality of the representation vectors should be more related to the downstream performance than the projection vectors, as verified in this subsection.

\begin{table}[h]
\caption{\textbf{Downstream error rate}. The correlation between the sample average of pair-wise cosine similarity $s$ and misclassification result $y$ as: $ \rho(s^{(r)}, y)$, $ \rho(s^{(z)}, y)$. The error rate, defined in \ref{eq_er}, stands for the two groups of error rate comparison, divided by 50\% percentile of the sample average of pair-wise cosine similarity $s$. 
(representation-based and projection-based error rates comparison are: $e_1^{(r)} | e_2^{(r)}$ and $e_1^{(z)} | e_2^{(z)}$. The SSL model is SimCLR trained for 200 epochs in Cifar10 with ResNet-18 as the encoder.}
\label{table:Correlation_label_accuracy}
\vskip 0.15in
\begin{center}
\begin{small}
\begin{sc}
\begin{tabular}{lcc}
\toprule
 Cifar10 &  Representation & Projection \\
\midrule
Correlation   &  0.1482 & 0.021 \\
Error rate split & 14.1\% $|$ 6.7\% &
11.1\% $|$ 9.7\% \\
\midrule
Cifar100 &  Representation & Projection \\
\midrule
Correlation  & 0.1746 & 0.0642 \\
Error rate split & 46.1\%$|$32.2\% & 42.1\%$|$36.2\% \\

\bottomrule
\end{tabular}
\end{sc}
\end{small}
\end{center}
\vskip -0.1in
\end{table}

We apply the sample average of pair-wise cosine similarity, defined as
\begin{align*}
    s^{(r)}_i = \frac{1}{n}\sum_{j!=i}^{j=n} r_i r_j,  s^{(z)}_i = \frac{1}{n}\sum_{j!=i}^{j=n} z_i z_j,
\end{align*}
where $r_i$ and $z_i$ stand for representation and projection vectors  of sample $i$, respectively. Define $y_i$ as the downstream misclassification of sample $i$, i.e., $y_i = 1$ represents misclassification in the downstream task while $y_i = 0$ otherwise. Define the correlation between the sample average of pair-wise cosine similarity $s^{(r)},s^{(z)}$ and misclassification result $y$ as $ \rho(s^{(r)}, y)$, $ \rho(s^{(z)}, y)$.
Intuitively, a sample with higher similarity with other samples indicates that this sample is difficult to be differentiated and thus more likely to be misclassified. 

\newcommand{\cG}{\mathcal{G}}

Table \ref{table:Correlation_label_accuracy} shows the correlations $ \rho(s^{(r)}, y)$, $ \rho(s^{(z)}, y)$ for the representation and projection vectors, respectively. It can be seen that the representation vectors are more related to the downstream task performance, as they possess a higher correlation. To further verify that the quality of the representation vectors can influence the downstream task performance more directly, we compute the error rate split for the representation and projection vectors. Specifically, we split the samples into two groups according to the pair-wise cosine similarity of representation vectors, i.e.,
\begin{align*}
    \cG_1^{(r)} = \{i: s^{(r)}_i > \mbox{median of $s^{(r)}$}\}, \cG_2^{(r)} = \{1,...,n\} \backslash \cG_1^{(r)}.
\end{align*}
The error rate for each group is defined as
\begin{align}\label{eq_er}
    e_j^{(r)} = \frac{1}{{\rm card}(\cG_j^{(r)})}\sum_{i\in \cG_j^{(r)}}y_i, \quad j=1,2.
\end{align}
The split and the error rates of the projection vectors can be obtained similarly. Intuitively, the error rate of $\cG_1^{(r)}$ (or $\cG_1^{(z)}$) is larger than that of $\cG_2^{(r)}$ (or $\cG_2^{(z)}$), since the samples in the former group are harder to be differentiated. The gap between the error rates of two groups should be large if the cosine similarity can precisely represent the difficulty of the downstream task. From Table \ref{table:Correlation_label_accuracy}, the gap in the representation vectors is much larger than that in the projection vectors, which reflects that the similarity within the representation vectors relates to the downstream task performance more closely.

\begin{figure*}
    \centering
    \includegraphics[width=0.8\textwidth]{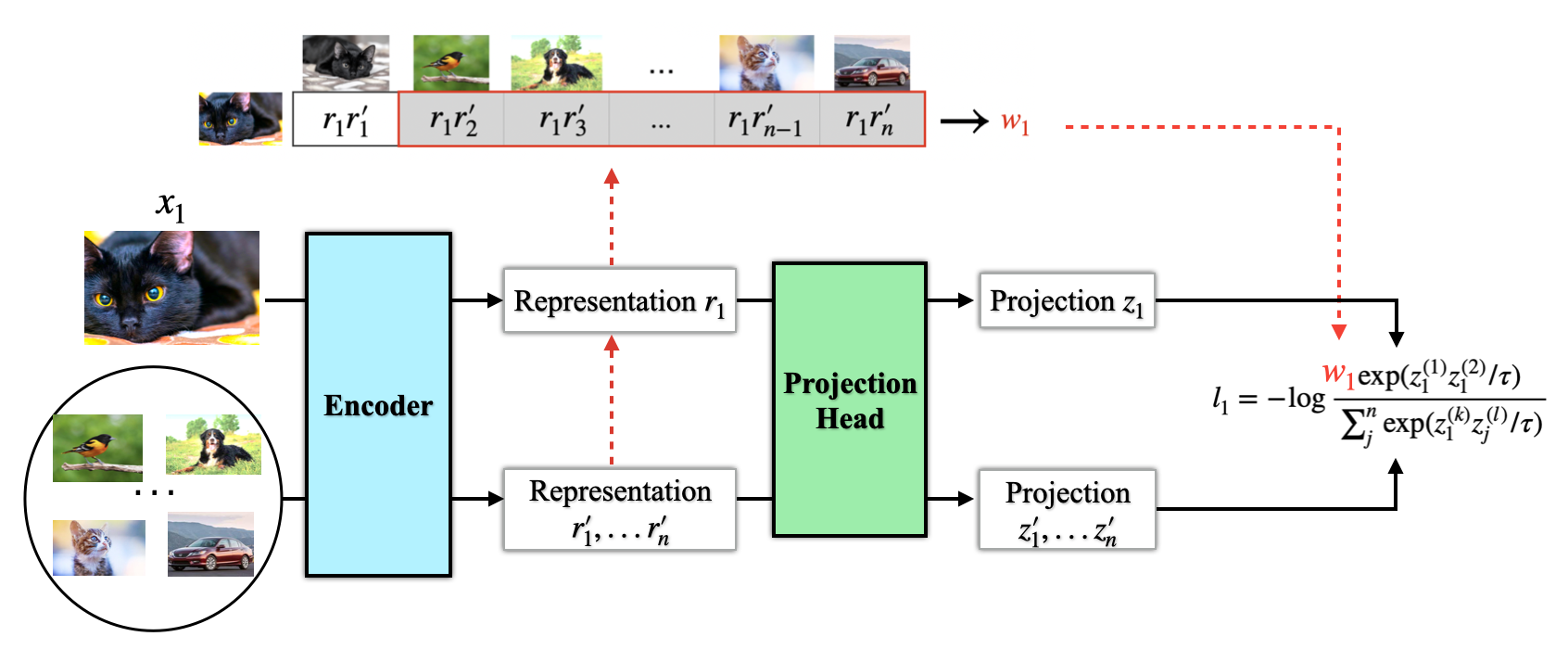}
    \caption{\textbf{RED-SimCLR demo}.}
    \label{fig:EvaCLR_algo_demo}
\end{figure*}

\section{Representation Evaluation Design in SSL}
The above results demonstrate that the representation vectors are more robust to the unseen augmentations, have a smaller entropy, and are more influential to the downstream task performance. These merits motivate us to propose our Representation Evaluation Design (RED), which allows us to utilize the advantages of the representation vectors directly.

The central idea of RED is to re-weight the positive alignment in the objective functions of SSL. As demonstrated in Section \ref{subsec_relation_task}, the representation vectors are more related to the downstream performance. Thus, the re-weight term actively guides the training for boosting downstream accuracy. Furthermore, we take advantage of the representation's superior augmentation robustness to adjust the alignment term of the projection vectors such that the RED-SSL equips features with stronger augmentation robustness. 

We propose to use the following weights for the alignment term of sample $i$:
\begin{align}\label{eqn:weight}
w_i = \frac{1}{ \text{percentile}_{j}(\exp(r_i r_j / \eta), k\%) },
\end{align}
where $\eta$ is the representation temperature parameter, and $k$ is the percentile parameter. In other words, $w_i$ is the reciprocal of the $k\%$ percentile of the exponential of pairwise representation product $\exp(r_i r_j / \eta)$. Instead of using the sample average, which encourages undesired uniformity and entropy increment in the representation level, the percentile can avoid encouraging uniformity and concentrate on alignment. 

Take SimCLR (constrastive) and SimSiam (non-constrastive) as example. With re-weighting shown in \eqref{eqn:weight}, the objective functions in SimCLR and SimSiam become
\begin{align}
L^{\text{SimCLR}} = & \sum_{i=1}^n -\log \frac{w_i \exp(z^{(1)}_i z^{(2)}_i / \tau )}{\sum_{j=1}^{N} \exp(z_i z_j / \tau) },\label{eqn:RepCLR_algorithm}\\
L^{\text{SimSiam}} = & \sum_{i=1}^n -\log w_i \exp(z^{(1)}_i z^{(2)}_i / \tau ),\label{eqn:PosCLR_algorithm}
\end{align}
respectively, where $\tau$ is a parameter called temperature parameter. Note that the individual loss for sample $i$ can be written as
\begin{align}
    l^{\text{SimCLR}}_{i} & = -\log w_i - \log \frac{ \exp(z^{(1)}_i z^{(2)}_i / \tau )}{\sum_{j=1}^{N} \exp(z_i z_j / \tau) },\label{eqn:clr}\\
      l^{\text{SimSiam}}_{i} & = -\log w_i - z^{(1)}_i z^{(2)}_i / \tau,\label{eqn:simsiam}
\end{align}
where the second terms in \eqref{eqn:clr} and \eqref{eqn:simsiam} are the original individual loss for sample $i$ in SimCLR and SimSiam, respectively. Therefore, RED-SSL essentially builds up a short-cut connection between the representation and projection vectors. During the \textit{batch} stochastic gradient descent, the gradients of $\log w_i$ can bypass the projection head and reach the representation vector directly, as illustrated in Figure \ref{fig:EvaCLR_algo_demo}: RED in SimCLR architecture. This connection also helps avoid the vanishing gradients of the representation vector.

\theoremstyle{remark}
\newtheorem{remark}{Remark}
\begin{remark}
Note that the \textit{batch} optimization is important in our RED. For example, if one use gradient descent, \eqref{eqn:PosCLR_algorithm} is the same as 
\begin{align*}
    L^{\text{SimSiam}} = & -\sum_{i=1}^n \log w_i - \sum_{i=1}^n z^{(1)}_i z^{(2)}_i / \tau,
\end{align*}
which is simply adding a term and is \textit{not} a re-weighting, thus cannot utilize the advantages of RED.
\end{remark}

\section{Experiments}
In this section, we compare the existing SSL models: SimCLR, MoCo-V2, and SimSiam, with our representation evaluation SSL (RED-SSL): RED-SimCLR, RED-MoCo-V2, and RED-SimSiam, respectively. We show RED can consistently boost downstream performance in Cifar10, Cifar100, ImageNet1000, and mixed-Gaussian data, for diverse SSL architectures. Similar to the skip connections' superiority in robustness to perturbations \cite{cazenavette2021architectural,reshniak2020robust}, we show that with RED, the SSL models can gain more robustness against unseen augmentations.

\subsection{Downstream task performance}
We set the downstream task as classification and compare the classification accuracy of SimCLR, MoCo-V2, SimSiam, and their corresponding representation evaluation models, RED-SimCLR, RED-MoCo-V2, and RED-SimSiam, respectively. 

\begin{table}[h]
\caption{\textbf{Downstream classification in Cifar10 and Cifar100 }. Comparison between the SimCLR, MoCo-V2, SimSiam, and their Representation evaluation (RED-) counterparts in Cifar10 and Cifar100, where the numbers in each column are the kNN$|$linear classifier accuracy.}
\label{table:200_downstream_knn_accuracy}
\vskip 0.15in
\begin{center}
\begin{small}
\begin{sc}
\begin{tabular}{lcc}
\toprule
    & Cifar10  & Cifar100 \\
    & kNN$|$linear  & kNN$|$linear \\
\midrule
SimCLR  & 81.4\%$|$83.0\% & 52.1\%$|$56.3\%    \\
RED-SimCLR &  84.6\%$|$86.4\% & 56.6\%$|$58.6\%   \\
MoCo-V2 & 83.2\%$|$85.1\% & 55.7\%$|$62.1\%    \\
RED-MoCo-V2 & 85.4\%$|$87.2\% & 58.5\%$|$63.5\%   \\
SimSiam	& 81.8\%$|$82.9\% & 50.7\%$|$52.0\%   \\
RED-SimSiam & 84.6\%$|$85.7\% & 51.7\%$|$52.7\%    \\
\bottomrule
\end{tabular}
\end{sc}
\end{small}
\end{center}
\vskip -0.1in
\end{table}

\begin{table}[h]
\caption{\textbf{Downstream classification in ImageNet1000 }.Comparison between the SimCLR, MoCo-V2, SimSiam, and their Representation evaluation (RED-) counterparts in ImageNet1000, where the numbers in each column are the kNN$|$linear classifier accuracy.}
\label{table:200_downstream_imagenet_knn_accuracy}
\vskip 0.15in
\begin{center}
\begin{small}
\begin{sc}
\begin{tabular}{lc}
\toprule
& ImageNet 1000 \\
     & kNN$|$linear\\
\midrule
MoCo-V2      &  44.9\%$|$67.5\%  \\
RED-MoCo-V2  &  54.4\%$|$68.0\%   \\
SimSiam     &  52.3\%$|$68.1\%   \\
RED-SimSiam &  53.4\%$|$68.3\%   \\
\bottomrule
\end{tabular}
\end{sc}
\end{small}
\end{center}
\vskip -0.1in
\end{table}

Table \ref{table:200_downstream_knn_accuracy} compares the classification accuracy of different SSL models in Cifar10 and Cifar100 data, where the downstream classifiers are k-nearest neighbors (kNN) and linear classifiers. For all models in Table \ref{table:200_downstream_knn_accuracy}, we set the batch size as 256, epochs as 200, and the encoder as ResNet-18. For RED-SimCLR and RED-MoCo-V2, the $\eta$ is 20.0, and $k\%$ is 95\%. For RED-SimSiam, $\eta$ is 100.0 and $k\%$ is 95\%. Table \ref{table:200_downstream_imagenet_knn_accuracy} compares the classification accuracy in ImageNet1000 data, where we choose 200 epochs of training for MoCo-V2, 100 epochs for training in SimSiam. The batch size is 256 for MoCo-V2 and 512 for SimSiam, and the encoder is ResNet-50. For RED-MoCo-V2 and RED-SimSiam, the $\eta$ is 100.0, and $k\%$ is 98\%.  We also compare the downstream task performance for the mixed-Gaussian simulated data, and the results are in Appendix \ref{appendix:mixed_gaussian}. The results demonstrate that RED can consistently improve the downstream task performance of the popular SSL models. Because of the simple structure of RED, it is adaptive to a wide range of SSL models, whether they are contrastive (e.g., SimCLR, MoCo-V2) or not (e.g., SimSiam).

\subsection{Robustness to (unseen) augmentations}

As discussed in Section \ref{subsec_robust}, the representation vectors are more robust to the unseen augmentations. Since the re-weights $w_i$ brings a short-cut connection between the representation and projection vectors, the RED-SSL models are supposed to be more robust to augmentations. In this subsection, we conduct experiments to show that with RED, the SSL models can be less affected by the (unseen) augmentations.

\begin{table}[h!]
\caption{\textbf{Classification accuracy on augmented data}.}
\label{table:compare_model_augmentation_accuracy}
\vskip 0.15in
\begin{center}
\begin{small}
\begin{sc}
\begin{tabular}{lcc}
\toprule
    & MoCo-V2  & RED-MoCo-V2 \\
\midrule
Train augmentation & 77.27\% & 79.58\% \\
Angle rotate      & 36.20\% & 38.61\% \\
Gaussian Blur  & 78.41\% & 80.60\% \\
Center Crop  &  15.41\%  & 18.83\% \\ 
Random Posterize  &  46.68\% & 47.57\%   \\
\bottomrule
\end{tabular}
\end{sc}
\end{small}
\end{center}
\vskip -0.1in
\end{table}

We adopt MoCo-V2 as our SSL model in this experiment. After training MoCo-V2 and RED-MoCo-V2, we apply (unseen) augmentation to the testing samples and evaluate the classification accuracy of the augmented testing data. The results are presented in Table \ref{table:compare_model_augmentation_accuracy}, showing how the augmentations affect classification accuracy. 
Compared with the original MoCo-V2, RED-MoCo-V2 consistently shows more robustness to diverse augmentations. Even for the train augmentation, RED enables the SSL model to gain additional robustness, which is preserved for unseen augmentations. Thus, our proposed representation evaluation design captures more semantic information that is augmentation invariant.

\subsection{Robustness to out-of-distribution data}

It has been shown that standard SSL methods, e.g., SimCLR, MoCo, etc., could suffer from out-of-distribution shift \cite{hu2022your}. This subsection indicates that RED can improve the out-of-distribution generalization. We train MoCo-V2 and RED-MoCo-V2 with Cifar10 (or Cifar100) and test the classification accuracy on Cifar100 (or Cifar10). The models are trained for 200 epochs with ResNet-18 as the encoder. The results are presented in Table \ref{table:200_ood_downstream_knn_linear}, exhibiting our RED enhancement for out-of-distribution generalization ability.

\begin{table}[h]
\caption{\textbf{Transfer learning}. kNN $|$ linear evaluation accuracy of out-of-distribution data. In the first and second rows, the models are trained with Cifar10, and in the third and fourth rows, the models are trained with Cifar100.}
\label{table:200_ood_downstream_knn_linear}
\vskip 0.15in
\begin{center}
\begin{small}
\begin{sc}
\begin{tabular}{lcccc}
\toprule
    & Cifar10  & Cifar100 \\
    & kNN$|$linear  & kNN$|$linear \\
\midrule
   MoCo-V2        & -  & 48.05\%$|$55.46\%    \\
   RED-MoCo-V2     &  - &  48.87\%$|$57.27\%  \\
   MoCo-V2        &  73.39\%$|$79.39\% & -    \\
   RED-MoCo-V2     &  74.29\%$|$79.86\% & -  \\
\bottomrule
\end{tabular}
\end{sc}
\end{small}
\end{center}
\vskip -0.1in
\end{table}

\subsection{Ablation study}
In this part, we conduct the ablation study on the two hyperparameters within the re-weighting term $w$ as in \eqref{eqn:weight} in RED: the percentile parameter $k$ and the representation temperature $\eta$. Note that percentile parameter $k$ stands for the $k\%$ percentile of the pair-wise representation product within the batch, and the representation temperature $\eta$ controls the magnitude of the representation evaluation impact. We compare the classification accuracy under different choices of $k$ and $\eta$. The models are RED-SimCLR, RED-MoCo-V2, and
RED-SimSiam, all are trained for 200 epochs. The downstream classifier is kNN, and the dataset is Cifar10.

The results are presented in Table \ref{table:ablation_percentile_temperature}. Table \ref{table:ablation_percentile_temperature} indicates that except for $\eta$ in SimSiam (which has been shown that it is sensitive to the model design \cite{li2022understanding}), SSL models with RED are relatively not sensitive to the choice of these two hyperparameters. Therefore, we can conclude that it is the design of RED that improves SSL models.

\begin{table}[h]
\caption{\textbf{Ablation study}. The ablation study of representation evaluation SSL models under different percentile $k\%$ and representation temperature $\eta$ parameters.}
\label{table:ablation_percentile_temperature}
\vskip 0.15in
\begin{center}
\begin{small}
\begin{sc}
\begin{tabular}{lcccc}
\toprule
  Percentile $k\%$  & 85\%  & 89\% & 95\% & 98\% \\
\midrule
RED-SimCLR  & 83.4\% & 84.6\% & 84.2\% & 83.9\% \\ 
RED-MoCo-V2  & 84.7\% & 85.4\% & 85.2\% & 85.2\% \\
RED-SimSiam & 83.8\% & 84.6\% & 83.8\% & 83.5\% \\
\midrule

Temperature $\eta$ & 10.0 & 20.0 & 50.0 & 100.0 \\
\midrule
RED-SimCLR  & 84.6\% & 84.5\% & 84.2\% &  84.0\% \\ 
RED-MoCoV2  & 84.8\% & 85.4\% & 84.8\% & 84.5\% \\
RED-SimSiam & 9.1\% &  22.9\% & 72.8\% & 84.6\% \\
\bottomrule
\end{tabular}
\end{sc}
\end{small}
\end{center}
\vskip -0.1in
\end{table}

\section{Conclusion}
This paper conducts a comprehensive analysis of the projection head design in SSL, uncovering that the projection head encourages uniformity, thus allowing the encoder to focus on boosting alignment. It explains that combining the encoder and projection head outperforms the individual encoder. With the projection head implicitly ensuring uniformity, it also explains the non-collapsed training of non-contrastive models such as SimSiam. 
The encoder \& projection head combination enables the encoder to spotlight boosting alignment without worrying about uniformity; thus, the representation vectors enjoy more robustness to augmentation, lower entropy, and better downstream task performance than the projection vectors. 

Based on these insights, we introduce the Representation Evaluation Design (RED), which is adaptive to diverse SSL models. We demonstrate that RED-SSL models outperform the corresponding baseline models in downstream task performance and exhibit more robustness to diverse augmentations and out-of-distribution data. Our research sheds light on the inner structure of SSL models and, hopefully, can motivate more research along this line.

\bibliographystyle{unsrt}  
\bibliography{contrastive_learning_latex}

\appendix

\appendix
\onecolumn
\newpage

\numberwithin{equation}{section}
\numberwithin{figure}{section}

\section{Uniformity and entropy}\label{appendix:Uniformity_entropy}

Under L2 normalization, the Euclidean distance is equivalent to the negative of cosine similarity, as shown in \eqref{eqn:euclidean_cosine}.  

\begin{equation}
\label{eqn:euclidean_cosine}
\begin{aligned}
\| x - y \|^2_2  & = x'x + y'y -2x'y \\ 
           & = \|x\|_2^2 + \|y\|_2^2 -2x'y  \\
         & = 2 - 2x'y \\
x'y & = 1 - \frac{1}{2} \| x - y \|^2_2 \\ 
 \end{aligned}
\end{equation}

The calculation of negative pairs of each sample $z_i$(denominator term of contrastive loss) can be viewed as a probability density estimation($\hat{p}_i$ is estimated based on von Mises-Fisher kernel $c_d(\kappa) e^{\kappa \mu'x}$) at the location $i$.   

\begin{equation}
\label{eqn:uniformity_probability}
\begin{aligned}
\log \frac{1}{\sum_j e^{z_i z_j / \tau}} & = 
-\log \frac{1}{n} \sum_j e^{z_i z_j / \tau} - \log n \\
& = -\log \frac{e^{1/\tau}}{n} \sum_j e^{-\|z_i - z_j \|_2^2 / 2\tau } - \log n \\ 
& = -\log \frac{1}{n} \sum_j e^{-\|z_i - z_j \|_2^2 / 2\tau } + \alpha_0 \\ 
& = -\log \hat{p}(z_i) + \alpha_0 \\
& = \log \frac{1}{\hat{p}(z_i)} + \alpha_0 \\
 \end{aligned}
\end{equation}

While for KL divergence between uniform distribution $U$ and estimated probability distribution $\hat{P}$ 
\eqref{eqn:kl_uniformity_probability}: it consists of the logarithm of reciprocal estimated probability, which corresponds to the result of uniformity in \eqref{eqn:uniformity_probability}. Thus the training of uniformity will lead the distribution of $z$  to approach the uniform distribution:

\begin{equation}
\label{eqn:kl_uniformity_probability}
\begin{aligned}
KL(U || \hat{P} ) & = \sum_i U(z_i) \log(\frac{U(z_i)}{\hat{p}(z_i)}) \\
& = \sum_i C \log(\frac{C}{\hat{p}(z_i)}) \\
& = C_0 \sum_i \log\frac{1}{\hat{p}(z_i)} + C_1 \\
& = C'_0 \sum_i \text{Uniformity}_{z_i} + C'_1 \\
 \end{aligned}
\end{equation}

\begin{equation}
\label{eqn:kl_uniformity_entropy}
\begin{aligned}
KL(\hat{P} || U ) & = \sum_i \hat{p}(z_i) \log(\frac{\hat{p}(z_i)}{u}) \\ 
& = \sum_i \hat{p}(z_i) \log(\hat{p}(z_i)) + c \\
& = - \text{Entropy}_{\hat{p}} + c \\
\end{aligned}
\end{equation}

The KL divergence to the uniform distribution can be transformed into a constant minus entropy of the sample distribution from \eqref{eqn:kl_uniformity_entropy}. Thus, the training in the projection-based uniformity will result in higher entropy of the projections.

\section{Mixed Gaussian}\label{appendix:mixed_gaussian} 
\begin{figure}
    \centering
    \includegraphics[width=0.3\textwidth]{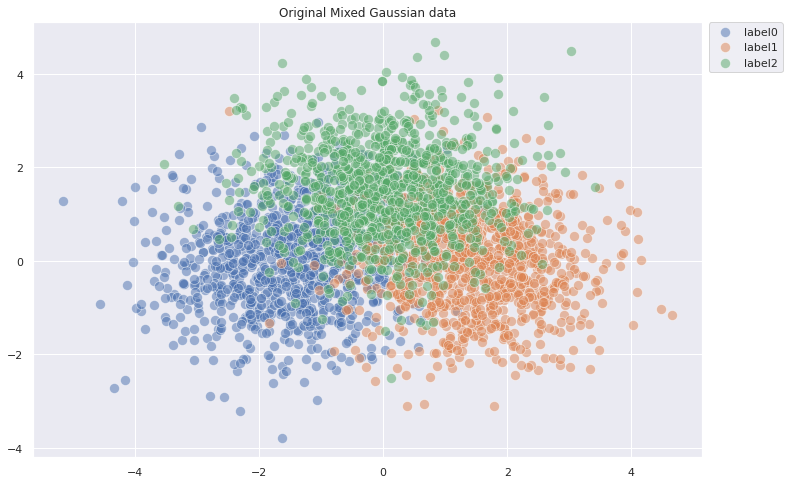}
    \caption{The simulated mixed Gaussian distributed samples}
    \label{fig:sim_original_samples}
\end{figure}

 The labeled data samples subjects to 2-dimensional Gaussian distribution: $(x, y=i): x \sim N(\mu_i, \sigma_i)$, shown in Figure \ref{fig:sim_original_samples}. Specifically, the 3 distributions are $N([0.5,0.7], \sigma), N([3.5,0.7], \sigma), N([2.0,3.3], \sigma)$, $\sigma$ is the identity matrix. The augmentation is defined as adding Gaussian noise to the sample: $ x^* = x + \epsilon, \epsilon \sim N(\mu_{\epsilon}, \sigma_{\epsilon})$, $\epsilon$ is 0.1. The model used for SSL consists of three blocks of MLP, each of which consists of a 10-dimensional linear layer and ReLU activation, and the last projection layer is 2-dimensional. The objective is infoNCE. The simplified model design shows us a clear view of the SSL process.

In Table  \ref{table:evaluation_gaussian_repr_proj}, we conduct the downstream kNN evaluation and record the positive alignment, uniformity, and entropy with the representation and projection vector, respectively. The downstream performance trained with representation is much better than that of projections. The positive alignment and the uniformity of representation are larger than the projection layer, indicating the representation layer is more augmentation robust while the projection layer is approaching closer to the uniform distribution on the hyper-sphere. 

\begin{figure}[ht]
\vskip 0.2in
\begin{center}
\centerline{\includegraphics[width=0.6\columnwidth]{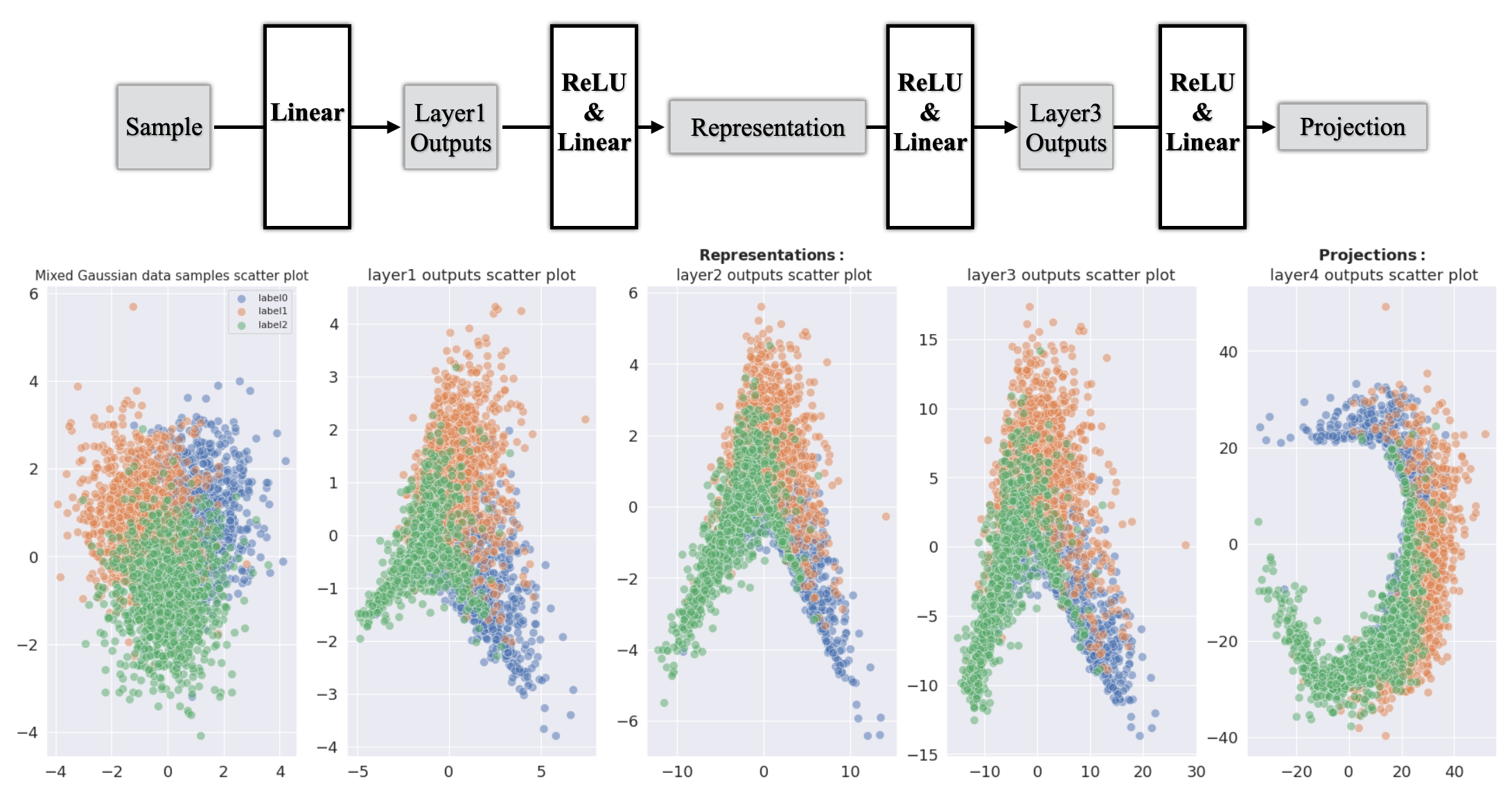}}
\caption{\textbf{Mixed-Gaussian data in layer-wise SSL output}. The distribution shift of the 2-dimensional mixed Gaussian data samples and the intermediate network layer outputs(reduced to 2 dimensions with PCA). The model consists of three blocks of linear layer and ReLU activation. }
\label{fig:sim_outputs_by_layer}
\end{center}
\vskip -0.2in
\end{figure}

Figure \ref{fig:sim_outputs_by_layer} depicts the dynamics of distribution shift of the original mixed Gaussian distributed enter through the network. The mixed Gaussian is gradually stretched along the axes in the beginning, and the scale becomes increasingly larger. For intermediate outputs of layer 1 to layer 3, we reduce the ten dimensions to 2 with PCA. We can see the angle between classes is smaller, representing the cosine distance decreases. At the last layer, the projection vectors are curled into a circle(the uniformity loss encourages samples to be evenly distributed on the surface of the hyper-sphere) at the cost of entropy increment. 

\begin{table}[t]
\caption{\textbf{Mixed-Gaussian data in SSL}. Evaluation of Representation and Projection in simulated mixed Gaussian data}
\label{table:evaluation_gaussian_repr_proj}
\vskip 0.15in
\begin{center}
\begin{small}
\begin{sc}
\begin{tabular}{lcc}
\toprule
        & Representation  & Projection \\
\midrule
KNN accuracy    &  70.5\% & 65.16\%   \\
Positive alignment  & 0.9721 & 0.9500 \\
Uniformity       & 0.7724 & 0.6643 \\
Entropy & 0.704 & 0.715 \\
\bottomrule
\end{tabular}
\end{sc}
\end{small}
\end{center}
\vskip -0.1in
\end{table}

\subsection{Mixed Gaussian simulated data}
The mixed Gaussian data is simulated with different labels subject to corresponding Gaussian distribution, as shown in Figure \ref{fig:sim_original_samples} of Appendix \ref{appendix:mixed_gaussian}. The contrastive model is simplified as four layers of a linear layer with ReLU activation function. Table  \ref{table:evaluation_gaussian_simulation} lists the comparison with estimated entropy and KNN accuracy.     

\begin{table}[h]
\caption{\textbf{Evaluation in mixed-Gaussian data}. Evaluation of Representation and Projection in simulated Mixed Gaussian data}
\label{table:evaluation_gaussian_simulation}
\vskip 0.15in
\begin{center}
\begin{small}
\begin{sc}
\begin{tabular}{lcc}
\toprule
     & InfoNCE  & RED-InfoNCE \\
\midrule
    KNN accuracy                          					&  71.09\% & 74.98\%   \\
    Entropy              								& 0.704 & 0.692   \\
\bottomrule
\end{tabular}
\end{sc}
\end{small}
\end{center}
\vskip -0.1in
\end{table}

\section{Network analysis}\label{appendix:network_analysis}
Figure \ref{fig:network_layer_std} of Appendix \ref{appendix:network_analysis} records the layer parameter standard deviation. Across the encoder part, we can see the volatility decreases gradually across the layers: indicating the convolution layer behaves more stable in extracting meaningful and robust features for positive pairs augmentation invariant. In contrast, stepping into the projection head layer, the network volatility spikes, indicating the training objectives shift into uniformity: mapping the data distribution into the uniformly distributed space. 

\begin{figure}[ht!]
\vskip 0.2in
\begin{center}
\centerline{\includegraphics[width=0.5\columnwidth]{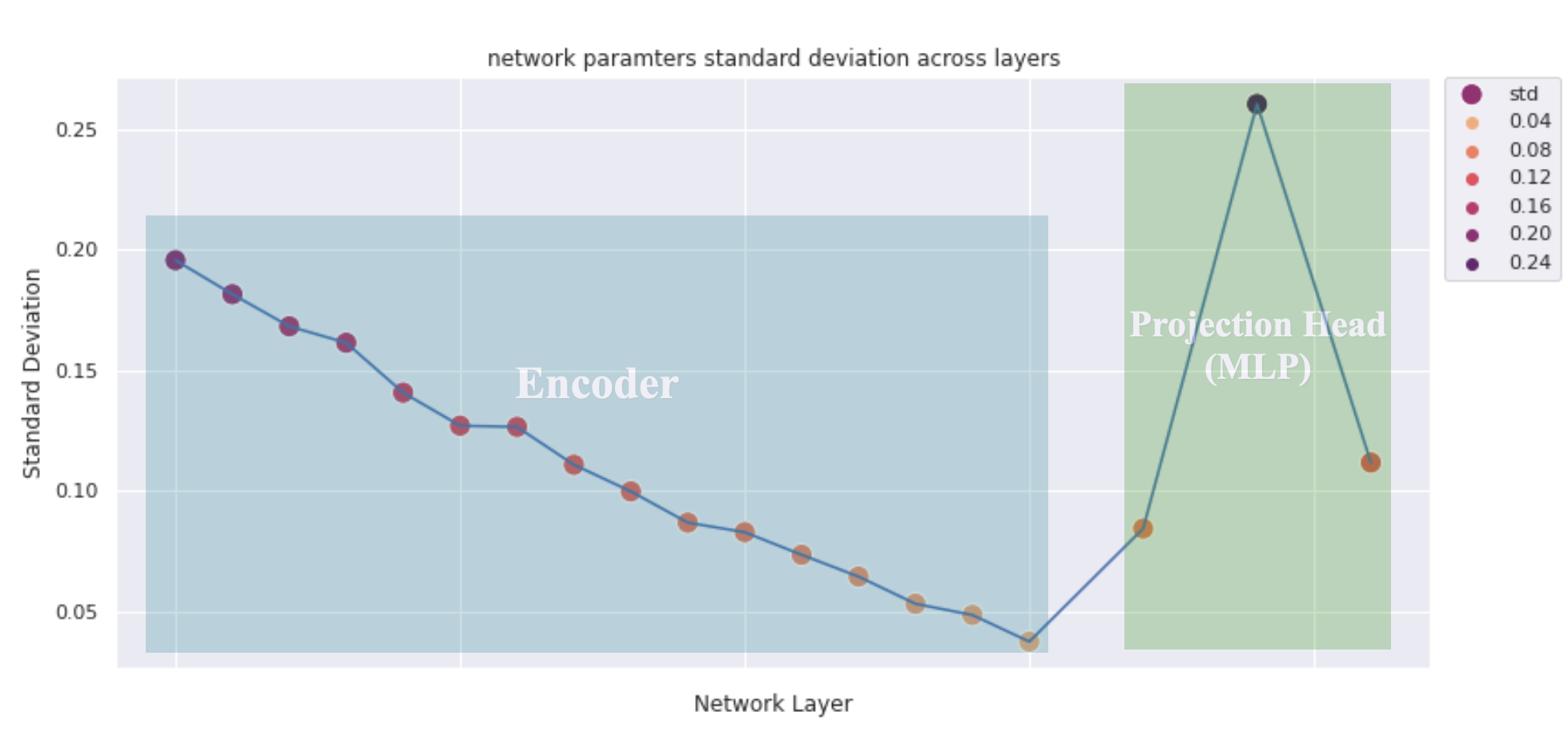}}
\caption{\textbf{Parameters standard deviation}. The intermediate layer parameters standard deviation of SimCLR network}
\label{fig:network_layer_std}
\end{center}
\vskip -0.2in
\end{figure}

\begin{figure}[ht!]
\vskip 0.2in
\begin{center}
\centerline{\includegraphics[width=0.5\columnwidth]{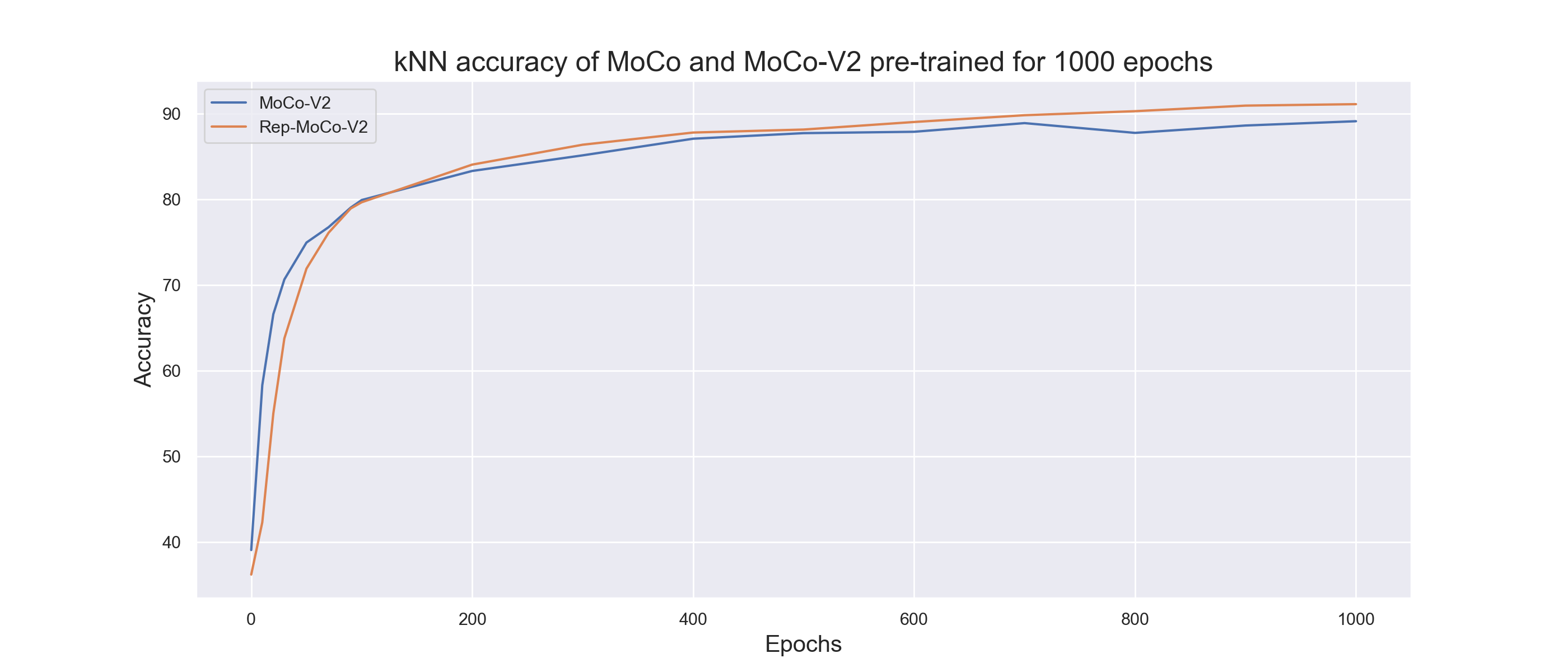}}
\caption{\textbf{Accuracy trajectory}.The kNN accuracy trajectory of MoCo-V2 and RED-MoCo-V2 trained for 1000 epochs}
\label{fig:1000_moco_knn_accuracy}
\end{center}
\vskip -0.2in
\end{figure}

\section{Ablation Study}
Figure \ref{fig:1000_moco_knn_accuracy} compares the MoCo-V2 and RED-MoCo-V2 downstream kNN accuracy trajectory of different contrastive models pre-trained for 1000 epochs.

\end{document}